\documentclass[sigconf]{acmart}

\pdfoutput=1

\copyrightyear{2024}
\acmYear{2024}
\setcopyright{rightsretained}
\acmConference[FORGE '24]{AI Foundation Models and Software Engineering}{April 14, 2024}{Lisbon, Portugal}
\acmBooktitle{AI Foundation Models and Software Engineering (FORGE '24), April 14, 2024, Lisbon, Portugal}\acmDOI{10.1145/3650105.3652289}
\acmISBN{979-8-4007-0609-7/24/04}

\usepackage[utf8]{inputenc}

\usepackage{xcolor}
\usepackage{algorithm}
\usepackage{algpseudocode}
\usepackage{calc}
\usepackage{numprint}
\usepackage{graphicx}
\usepackage{svg}
\usepackage[frozencache,cachedir=minted-output]{minted}
\usepackage{booktabs}
\usepackage{multicol}
\usepackage{multirow}
\usepackage{arydshln}
\usepackage{bbding}

\graphicspath{{./images/}}

\newcommand{\temp}[1]{{\color{orange} #1}}
\renewcommand{\temp}[1]{}

\newcommand{\optional}[1]{{\color{blue} #1}}
\renewcommand{\optional}[1]{}

\AtBeginDocument{%
  }

\begin{document}

\title{Investigating the Performance of Language Models 
for Completing Code in Functional Programming Languages: a Haskell Case Study}

\author{Tim van Dam}
\email{t.o.vandam@student.tudelft.nl}
\orcid{0009-0009-3659-7068}
\affiliation{%
  \institution{Delft University of Technology}
  \country{Delft, Netherlands}
}

\author{Frank van der Heijden}
\email{f.n.m.vanderheijden@student.tudelft.nl}
\orcid{0009-0006-5922-1221}
\affiliation{%
  \institution{Delft University of Technology}
  \country{Delft, Netherlands}
}

\author{Philippe de Bekker}
\email{p.m.debekker@student.tudelft.nl}
\orcid{0000-0002-3722-5428}
\affiliation{%
  \institution{Delft University of Technology}
  \country{Delft, Netherlands}
}

\author{Berend Nieuwschepen}
\email{b.r.nieuwschepen@student.tudelft.nl}
\orcid{0009-0008-8146-550X}
\affiliation{%
  \institution{Delft University of Technology}
  \country{Delft, Netherlands}
}

\author{Marc Otten}
\email{m.j.c.otten@student.tudelft.nl}
\orcid{0009-0006-6525-8127}
\affiliation{%
  \institution{Delft University of Technology}
  \country{Delft, Netherlands}
}

\author{Maliheh Izadi}
\email{m.izadi@tudelft.nl}
\orcid{0000-0001-5093-5523}
\affiliation{%
  \institution{Delft University of Technology}
  \country{Delft, Netherlands}
}

\begin{abstract}
Language model-based code completion models have quickly grown in use, helping thousands of developers write code in many different programming languages.
However, research on code completion models typically focuses on imperative languages such as Python and JavaScript, which results in a lack of representation for functional programming languages.
Consequently, these models often perform poorly on functional languages such as Haskell.
To investigate whether this can be alleviated, we evaluate the performance of two language models for code, CodeGPT and UniXcoder, on the functional programming language Haskell.
We fine-tune and evaluate the models on Haskell functions sourced from a publicly accessible Haskell dataset on HuggingFace.
Additionally, we manually evaluate the models using our novel translated HumanEval dataset.
Our automatic evaluation shows that knowledge of imperative programming languages in the pre-training of LLMs may not transfer well to functional languages, but that code completion on functional languages is feasible.
Consequently, this shows the need for more high-quality Haskell datasets.
A manual evaluation on HumanEval-Haskell indicates CodeGPT frequently generates empty predictions and extra comments, while UniXcoder more often produces incomplete or incorrect predictions.
Finally, we release HumanEval-Haskell, along with the fine-tuned models and all code required to reproduce our experiments on GitHub~\cite{authors2024replication}.

\end{abstract}


\maketitle

\section{Introduction}
Automatic code completion has already proven to be an asset for programmers~\cite{izadi2022codefill,MoradiDakhel2023, svyatkovskiy2020intellicode,izadi2024language,sergeyuk2024ide}, 
with many Large Language Models (LLMs) accurately predicting developers' intent given sufficient expert context.
GitHub Copilot has been very successful~\cite{nguyen2022empirical,vaithilingam2022expectation}, and case studies on automatic code completion such as IntelliCode Compose~\cite{svyatkovskiy2020intellicode}, Tabnine, and Amazon CodeWhispherer~\cite{yeticstiren2023evaluating}, 
prove that developers benefit from these completions.

Most of the training data for LLMs for code consists of imperative and Object-Oriented Programming (OOP) languages, with the most prominent programming languages being Python, JavaScript, and Java~\cite{katzy2023impact}.
While such languages have embraced many functional programming concepts and features over the past years, they are still mainly object-oriented.
In contrast to these popular languages, there is little research on code completion for strongly typed functional languages with advanced type class techniques.
As a result, most datasets contain little to no functional code, leading to poor performance in functional languages such as Haskell. These languages are, however, still frequently used in practice, albeit not nearly as often as imperative languages.
In this study, we aim to fill this gap by evaluating the performance of line completion in the Haskell programming language.
Haskell in particular is interesting, as the syntax and language constructs are more concise and considered more difficult by many programmers.
However, whether this also applies to language models has not been investigated.

We fine-tune two multilingual pre-trained code completion models, UniXcoder~\cite{guo2022unixcoder} and Code\-GPT~\cite{lu2021codexglue}, using Haskell functions from the Blastwind dataset, 
a publicly accessible dataset containing Haskell function implementations\footnote{\url{https://huggingface.co/datasets/blastwind/github-code-haskell-function}}.
We evaluate the models automatically using the aforementioned dataset and introduce a new dataset -- a version of HumanEval~\cite{chen2021evaluating} manually translated to Haskell -- that we use to manually evaluate the models.
For this, we translate \numprint{164} Python functions into their Haskell equivalent and introduce points of interest for our manual evaluation in the Haskell sources for the models to perform code completion on.

Our results show that models fine-tuned on Blastwind outperform base models by a significant margin, indicating that knowledge of a broad array of imperative programming languages does not transfer well to functional languages.
For our newly created HumanEval dataset translated into Haskell, the average performance, as measured by Exact Match (EM) and Edit Similarity (ES), indicates lower performance than on the Blastwind dataset.
Our manual evaluation of the performance of the models on the HumanEval dataset indicates several key differentiators in the behavior of the fine-tuned CodeGPT and UniXcoder.
CodeGPT often generates empty predictions and unnecessary comments, while UniXcoder has more \texttt{incomplete}, \texttt{wrong syntax}, or \texttt{undefined} predictions. This indicates that CodeGPT is more `careful' when predicting, while UniXcoder prefers to predict incorrect output.
Additionally, our manual evaluation shows no common pitfalls for Haskell code completion based on the distribution of annotations. Both models lack the performance to excel in any particular category, indicating a need for better datasets and models.
In short, CodeGPT was found to be relatively reliable and accurate than UniXcoder, resulting in CodeGPT being the safer choice for practical usage.
However, the manual analysis indicates that our fine-tuned models are not reliable enough to conclude that any particular aspect of functional programming is more difficult for code completion models.

\noindent To summarize, this paper highlights the potential for LLMs to comprehend functional concepts and languages, as shown by the performance increase resulting from fine-tuning. Additionally, the large performance gap between base models and fine-tuned models indicates that code completion models that are proficient in a wide array of imperative programming languages do not necessarily transfer this proficiency to functional languages.
%
Based on the lack of performance in all annotated categories of the manual evaluation, no specific pitfalls are detected for Haskell code completion. Instead, the results indicate a high demand for better overall language support.
This paper aims to accelerate this process by releasing a manually translated HumanEval dataset, originally written in Python, for Haskell, forming a great foundation for enhancing the performance of functional programming languages in further research.

The contributions of this paper are as follows:
\begin{itemize}
    \item An extensive evaluation of the performance of pre-trained and fine-tuned code completion models on Haskell;
    
    \item A comprehensive assessment on the primary pitfalls of code completion when applied to Haskell;
    
    \item Publicly available manual translation of HumanEval to Haskell for future evaluation, including marked locations of interest for code completion;
    
    \item Publicly available source code and fine-tuned code completion models for Haskell~\cite{authors2024replication}.
    
\end{itemize}

\section{Motivation}
Our work investigates the performance of LLM-based code completion, and in particular, line completion.
We perform our evaluation using the Haskell programming language, as we find Haskell (and functional languages in general) to be underrepresented within the realm of Natural Language Processing (NLP) for code.
Nearly all code completion models in the literature are trained on non-functional languages such as Python, Java, JavaScript, Go, and Ruby~\cite{guo2022unixcoder,lu2021codexglue,feng2020codebert,fried2022incoder,guo2021graphcodebert,wang2021code,svyatkovskiy2020intellicode}.
Our primary aim is to investigate how well code completion models work on Haskell.
Being a functional language, Haskell is syntactically and conceptually very different from the imperative programming languages that code completion models are typically trained on.
Some languages such as Java and JavaScript contain functional paradigms, but this is not as close to a purely functional language such as Haskell.
A notable difference, for instance, is the lack of control-flow statements such as \texttt{for} and \texttt{while} loops.
Hence, to evaluate the usefulness of code completion models on functional languages, we fine-tune code completion models on Haskell and evaluate their performance relative to other programming languages.
Concretely, we investigate code completion performance on a corpus of permissively licensed Haskell functions sourced from HuggingFace.\footnote{\url{https://huggingface.co/datasets/blastwind/github-code-haskell-function}}
Additionally, we manually evaluate code completions on a newly created dataset from HumanEval containing $164$ problems, which have been translated from the original Python source into Haskell function implementations for this purpose.

\section{Related Works}
This section will discuss the related works in the field of NLP with regard to the task of code completion. First, we will briefly discuss the model architecture, then we will look into some examples of models that are currently being used. Lastly, we discuss evaluations of models and empirical analysis results.

\subsection{Transformer Architecture}
LLMs for code completion take code as input and predict a segment of new code related to the input.
This can be done using Causal Language Modeling in conjunction with specialized ways of masking inputs to facilitate auto-regressive generation or infilling.
State-of-the-art models that are currently used for code completion use the transformer architecture~\cite{vaswani2017attention}.
By making use of attention layers, the transformer models allow for a better understanding of the semantic and syntactic structure of code. This is especially important in code, as code has very strict syntax, unlike natural language.
There are three main distinctions in the transformer model:

\subsubsection{Encoder-only models}
This type of architecture is commonly used for masked token prediction and classification.
At each stage, the attention layers can access all the words in the initial sentence (i.e. bi-directional attention).
The BERT~\citep{devlin2018bert} family of models are implementations of this architecture.
It has been studied widely, and has also been adapted for code~\citep{feng2020codebert,ciniselli2021empiricalTSE,Karmakar2021,Mashhadi2021}.


\subsubsection{Decoder-only models}
The decoder architecture is commonly used for code-completion and text generation.
Decoder-only tokens have access to the weights of the tokens that come before itself (i.e. the left context).
A family that implements this architecture is GPT~\cite{Radford2018gpt1}.
Large types of these models have been shown to be effective for dialogue~\cite{Pawel2019} and few-shot learning~\cite{Liu2021}.
Newer variations in this family are GPT-2~\cite{radford2019gpt2}, GPT-3~\cite{brown2020gpt3}, and GPT-4~\cite{OpenAI2023}.
They have been widely used in the industry~\cite{OpenAI2022}, and evaluated~\cite{Bubeck2023}.


\subsubsection{Encoder-Decoder models}
Encoder-Decoder models are a combination of the previous two architectures and have been popular in RNNs as ``seq2seq'', before the Transformer architecture~\cite{Sutskever2014}, but even in recent years it has still shown to be viable and competitive~\cite{Soltan2022}. The Encoder-Decoder architecture excels in machine translation, as the Encoder processes the source language and the Decoder processes this to the target language.

\subsection{Code Completion}
Automatic code completion can be used to predict single tokens, complete lines of code, or even complete functions.
Recently there has been a surge in developers using better code completion tools, such as Copilot, which has shown improvements in efficiency by using the completions and then slightly tweaking them~\cite{MoradiDakhel2023}.
Most code completion models follow the decoder-only architecture
and are trained using Unidirectional Language Modeling (ULM) on datasets containing natural language and code in many different programming languages.
For instance, GPT-C~\cite{svyatkovskiy2020intellicode} and CodeGPT~\cite{lu2021codexglue} are code completion models based on GPT-2~\cite{radford2019gpt2} that use unidirectional language modeling to learn from examples.
Similarly, Codex~\cite{chen2021evaluating} is a GPT-3-based model~\cite{brown2020gpt3} that can predict the next token(s) based on the current code.
While most models can only use the \textit{left context}, i.e., the source code that is to the left of the cursor, models like InCoder use Causal Masking to be able to handle code from the left and to the right of the cursor despite being auto-regressive~\cite{fried2022incoder,Aghajanyan2022}.
SantaCoder~\cite{allal2023santacoder} and StarCoder~\cite{li2023starcoder} similarly are trained to \textit{infill} code fragments between left and right contexts.
As an alternative technique to improve performance, numerous models use
modalities aside from source code, such as Abstract Syntax Trees (ASTs).
CodeFill~\cite{izadi2022codefill}, for instance, uses source code along with AST token types to enhance the input, leading to improved performance.
UniXcoder~\cite{guo2022unixcoder} similarly uses ASTs for a range of code generation and understanding tasks, however, does not benefit from ASTs when used for code completion.
While code completion is a widely studied topic,
surprisingly little research has been conducted on applying code completion
to functional languages.
A recent advancement has been made with datasets for OCaml, Racket and Lua~\cite{cassano2024knowledge}, where high-quality datasets were made for those underrepresented languages in LLMs.
However, almost none of the aforementioned code completion models use functional languages in their base-model training data, as shown in \autoref{tab:modellanguages}.
This indicates a clear need for more awareness.

\begin{table}[h]
    \caption{Code completion models and the languages they are trained on.}
    \label{tab:modellanguages}
    \centering
    \begin{tabular}{lp{6cm}}
        \toprule
        \textbf{Model}              & \textbf{Language(s)} \\
        \midrule
        GPT-C~\cite{svyatkovskiy2020intellicode} & C\#, Python, JavaScript, TypeScript  \\
        CodeGPT~\cite{lu2021codexglue} & Python, Java \\
        Codex~\cite{chen2021evaluating} & Python \\
        InCoder~\cite{fried2022incoder} & 28 PLs, no significant amount of functional code \\
        SantaCoder~\cite{allal2023santacoder} & Python, Java, JavaScript \\
        CodeFill & Python~\cite{izadi2022codefill} \\
        UniXcoder\cite{guo2022unixcoder} & CodeSearchNet~\cite{husain2019codesearchnet}, containing Python, Ruby, Java, JavaScript, PHP, and Go  \\
        StarCoder & 358 PLs, \textbf{0.291\% Haskell} \\ 
        \bottomrule
    \end{tabular}
\end{table}

Only StarCoder has Haskell in its training data~\cite{li2023starcoder}.
StarCoder uses the Stack dataset~\cite{kocetkov2022stack}, which contains permissively licensed open-source code sourced from GitHub, including 358 unique programming languages.
However, Haskell accounts for only 0.291\% of StarCoder's training data, and no investigation into the performance of Haskell was performed.

\section{Approach}
\label{sec:Approach}
\begin{figure*}[tb]
    \centering
    
    \includegraphics[width=1\textwidth]{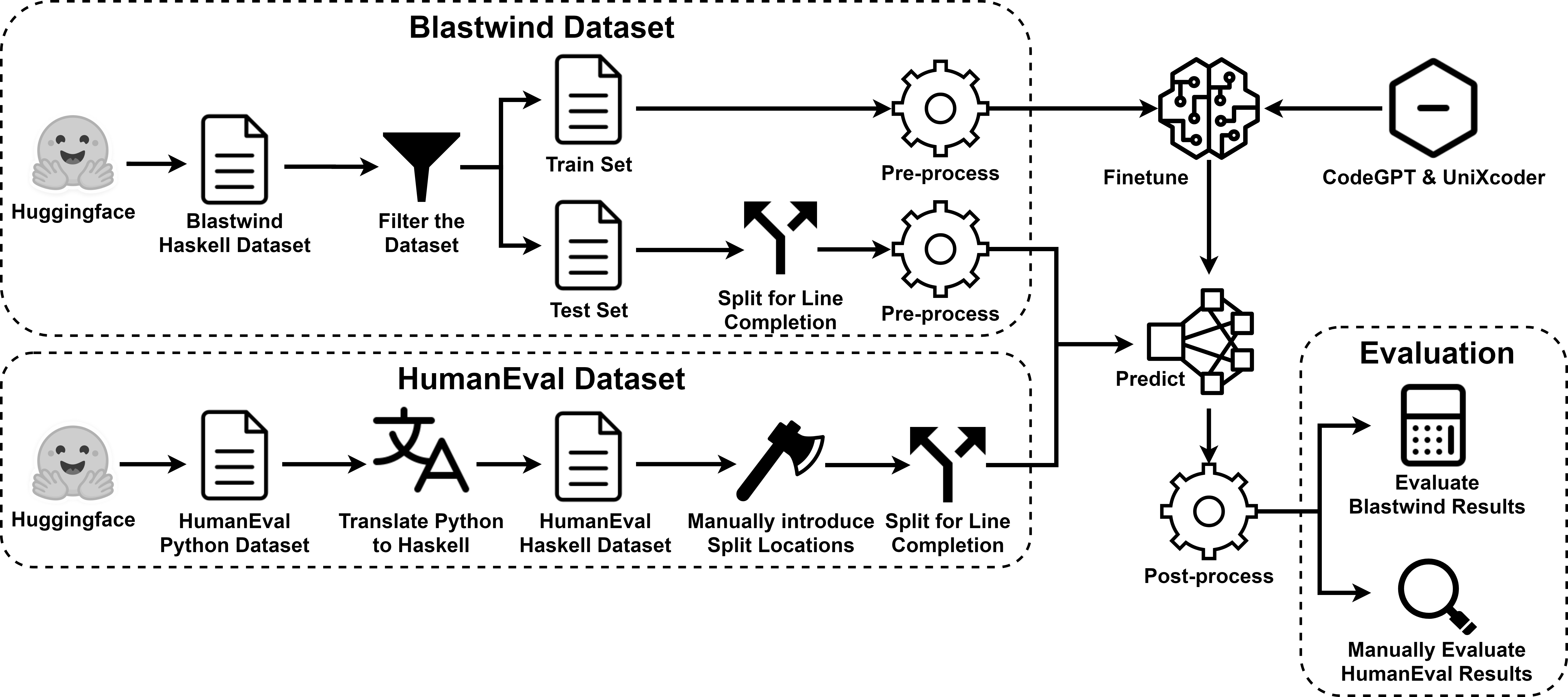}
    \caption{Approach pipeline.}
    \label{fig:approach_fig}
\end{figure*}

Our approach can be divided into three distinct steps: \textit{dataset creation}, \textit{fine-tuning}, and \textit{evaluation}.
During dataset creation, we process and split our data for use by the selected models.
Next, during fine-tuning, we train the models to perform line completion on Haskell code, using our train set.
Finally, during evaluation, we run the fine-tuned models and base models against our test sets.
An overview of the full approach pipeline is displayed in \autoref{fig:approach_fig}.

\subsection{Dataset Creation}
During the dataset creation phase, we organize datasets for training and evaluation.
We utilize two distinct datasets to achieve this.
The Blastwind dataset, sourced from HuggingFace, consists of publicly available Haskell function implementations.
This dataset serves as the foundation for the training phase.
The second dataset, which we created ourselves, 
involves translating HumanEval from Python to Haskell.
This self-created dataset is solely used for evaluation, which includes a manual evaluation where our objective is to identify the primary pitfalls to Haskell code completion.

\subsubsection{Blastwind Dataset}

\begin{figure}[tb]
    \begin{minted}[frame=single,linenos,breaklines,breaksymbol={},numbersep=1mm]{haskell}
-- | Create a pair generator.
pairOf :: Applicative m => m a -> m (a, a)
pairOf m = (,) <$> m <*> m
    \end{minted}
    \caption{Blastwind sample.}
    \label{fig:dataset_1_sample}
\end{figure}

This dataset contains Haskell function implementations. This dataset was selected as it was one of the very few available Haskell datasets and was large enough to be usable.
An example input from this dataset is shown in \autoref{fig:dataset_1_sample}.

Before using this dataset, we apply various processing steps to remove low-quality samples.
Initially, the dataset has \numprint{2287379} samples.
First, we apply a basic filtering step to remove low-quality samples, ensuring sufficient context for the model.
We use the following rules:
\begin{itemize}
    \item The code must have a comment;
    \item The code must have a function signature;
    \item Parsing must not result in any AST errors;
    \item The code must be at least two lines of code (excluding comments);
    \item The code must be at least 75 characters long (excluding comments).
\end{itemize}
This step removes 83.78\% of samples.
This suggests a lot of the samples in this dataset were uninformative samples.
Next, we deduplicate the remaining samples as duplicate samples can lead to data leakage between train and test sets.
Such data leakage and consequently result in over-inflated performance scores~\cite{allamanis2019adverse}.
Near duplicate deduplication is an expensive $\mathcal{O}(n^2)$ operation, especially considering the size of this dataset.
Hence, we are limited to performing exact-match deduplication.
This is relatively efficient through the use of hash-sets, which allow for duplicate checking in amortized $\mathcal{O}(1)$ time per sample.
Applying near-duplicate deduplication requires tokenizing all samples and cross-comparing all tokens for all sample pairs, which adds another layer of complexity of at least $\mathcal{O}(n + m)$ time depending on the way the tokens are compared.
Despite its limitations, exact-match deduplication removes 25.25\% of the remaining samples.

Our final dataset contains \numprint{277337} samples, which is 12.12\% of the original dataset size.
The processed dataset is subsequently split into a train and a test set using an 80\%-20\% split.
During splitting, we ensure that functions from the same repositories are in the same set, which leads to a 72.81\%-27.19\% split when counting the number of functions per split, corresponding to \numprint{201921} train samples and \numprint{75416} test samples.
This splitting approach serves to further prevent data leakage; a file-based split could lead to situations where a function from the train set is called in a function in the test set.
The samples are still full function implementations at this point, so we must convert the test samples into input-output pairs to evaluate our models.
\autoref{alg:io} describes the way these input-output pairs are created.
In short, whitespace characters in the sample are candidate split-points if it is not in a comment and has sufficient preceding and succeeding tokens.
We choose where to split by pseudo-randomly selecting from the candidate split-points.
Note that this splitting method may not accurately capture the use of line completion in practice.
One could argue that invoking code completion after specific \textit{trigger-points} such as \texttt{=}, \texttt{(}, and \texttt{.} may be more realistic.
However, it has been shown that even this strategy also does not strongly match programmer behavior in practice \cite{izadi2024language}.

\begin{algorithm}
\caption{Generate Input-Output Pair}
\label{alg:io}
\begin{algorithmic}[1]
\State $\text{whitespaceIndices} \gets \varnothing$
\For{character, index \textbf{in} code}\\
    \textbf{\hspace{\widthof{M}}if}
        \begin{align*}
            &\text{\hspace{\widthof{M}}character is a whitespace}\ \land \\
            &\text{\hspace{\widthof{M}}previous character is not a whitespace}\ \land \\
            &\text{\hspace{\widthof{M}}has at least five preceding tokens}\ \land \\
            &\text{\hspace{\widthof{M}}has at least one preceding token on the same line} \ \land \\
            &\text{\hspace{\widthof{M}}has at least two following tokens on the same line} \ \land \\
            &\text{\hspace{\widthof{M}}is not on a line starting with \texttt{"-{}-"}}\ \land \\
            &\text{\hspace{\widthof{M}}is not in a multi-line comment block}
        \end{align*}
    \textbf{\hspace{\widthof{M}}then}
        \State $\text{whitespaceIndices} \gets \text{whitespaceIndices} \cup \left\{\text{index}\right\}$ \\
    \textbf{\hspace{\widthof{M}}end if}
\EndFor
\State $\text{chosenIndex} \gets \text{pseudoRandomSelect(whitespaceIndices, seed)}$
\State $\text{Input} \gets \text{code up to chosenIndex}$
\State $\text{Output} \gets \text{code after chosenIndex up to first line-break}$
\State \Return (Input, Output)
\end{algorithmic}
\end{algorithm}

\subsubsection{HumanEval-Haskell Dataset} \label{approach:dataset-creation:dataset-2}

\begin{figure}[tb]
    \begin{minted}[frame=single,linenos,breaklines,breaksymbol={},numbersep=1mm]{haskell}
-- Check if in given list of numbers, are any two numbers closer to each other than
-- given threshold.
-- >>> has_close_elements [1.0, 2.0, 3.0] 0.5
-- False
-- >>> has_close_elements [1.0, 2.8, 3.0, 4.0, 5.0, 2.0] 0.3
-- True
has_close_elements :: [Float] -> Float -> Bool
has_close_elements numbers threshold = any (\(x, y) -> abs (x - y) < threshold) [(x,y) | x <- numbers, y <- numbers, x /= y]
    \end{minted}
    \caption{HumanEval-Haskell sample.}
    \label{fig:dataset_2_sample}
\end{figure}

We manually create a new dataset by translating HumanEval\footnote{\url{https://huggingface.co/datasets/openai_humaneval}} from Python to Haskell, a dataset that is often used for evaluating model performance.
An example of a translated function is shown in \autoref{fig:dataset_2_sample}.
The aim of this dataset is to be able to determine the most common mistakes by the selected models when applied to Haskell code.
As Haskell is such a unique language, we expect the models to encounter different pitfalls than in imperative, non-functional languages.
The translated dataset contains code that follows an identical structure to the original Python code: the code consists of an instructive comment including test input and output, a function signature, and an implementation. While translating, the functional concepts of Haskell, such as pattern matching, monads, and working without side effects, are properly implemented.
Additionally, any Python syntax embedded within the instructive comment is translated to Haskell syntax -- e.g., \texttt{my\_fn([1, 2, 3])} is translated to \texttt{my\_fn [1,2,3]}.
As our aim is to determine the most common pitfalls using line completion, we create equivalent Haskell functions from the one in the original HumanEval and run the new Haskell dataset on these.
Each author translated the same number of HumanEval test cases, after which each translated Haskell function was reviewed by two different authors to resolve any issues.

Next, points of interest in the code are manually introduced and marked using a special symbol.
We use the following points of interest, as they mark logical invocation points of code completion based on developer interest:

\begin{itemize}
    \item statements such as \texttt{if}/\texttt{then}/\texttt{else}, generators and guards;
    \item assignment operators such as \texttt{=}, \texttt{<-} and \texttt{->};
    \item logical operators such as \texttt{\&\&}, \texttt{||}, \texttt{==}, \texttt{>}, and more;
    \item arithmetic operators such as \texttt{/}, \texttt{*}, and more.
\end{itemize}

\noindent We never place these symbols at the very beginning or end of lines. Multiple splits can be introduced in a single Haskell function.

\subsection{Fine-tuning}
During the fine-tuning phase, the chosen code completion models are trained on the train set of the Blastwind dataset.
The models are fine-tuned to perform line completion.
All newline characters (\texttt{\textbackslash n}) in the training data are replaced by end-of-line tokens (\textbf{<EOL>}).
These tokens can be detected during inference and will cause the model to end its prediction loop.

\subsubsection{Selected Models}

We perform our experiments using two pre-trained decoder-only models for code completion.

First, \textit{UniXcoder}~\cite{guo2022unixcoder} is a unified pre-trained model
that can facilitate numerous code understanding and generation tasks, leveraging code comments and ASTs.
UniXcoder was trained using Masked Language Modeling~\cite{devlin2018bert,Baevski2019} and Denoising~\cite{raffel2019t5} to facilitate code understanding, and trained using Unidirectional Language Modeling~\cite{Radford2018gpt1} for auto-regressive tasks such as code completion.
UniXcoder was first trained on the C4 dataset~\cite{raffel2019t5} for English understanding,
after which it was trained on CodeSearchNet~\cite{husain2019codesearchnet}, which includes function implementations (including comments) written in six programming languages: Python, Java, JavaScript, Go, PHP, and Ruby.
While UniXcoder is able to also utilize an encoder-only or encoder-decoder mode, we have chosen specifically for decoder-only as this mode suits code completion best.

Second, \textit{CodeGPT}~\cite{lu2021codexglue} is a GPT-2-based~\cite{radford2019gpt2} model for code completion.
\citeauthor{lu2021codexglue} created four different versions of CodeGPT: for both Python and Java the authors create one version that uses the GPT-2 tokenizer and GPT-2 weights as a starting checkpoint, and one version that uses a newly trained tokenizer and starts with random weights.
GPT-2 was trained on the WebText dataset, which~\citeauthor{radford2019gpt2} introduced as a source of high-quality online text.
For our experiments, we use a version of CodeGPT that uses the GPT-2 tokenizer and checkpoint and is further trained on CodeSearchNet~\cite{husain2019codesearchnet}.\footnote{\url{https://huggingface.co/AISE-TUDelft/CodeGPT-Multilingual}}
This is similar to UniXcoder: both have been trained using a broad range of languages, which may be helpful when predicting new languages.
Both models have been trained on imperative languages, which could transfer to improved performance in functional settings.

\optional{
We train two variants for each chosen model: one trained on plain Haskell code, and one with the additional next-token-type information.
The latter is achieved by expanding the model input to accept a one-hot encoded vector where each element represents a different token type.
}

\subsection{Evaluation}
The evaluation phase applies the base models and fine-tuned models on the test set of Blastwind and on our newly created HumanEval-Haskell dataset.

\subsubsection{Post-Processing}
Before evaluation can begin the predictions need to be post processed.
This is to ensure the evaluations are accurate and fair.
The main steps in this process are stripping trailing spaces and newlines.
Additionally the predictions are normalised so the exact match can accurately be calculated.

\subsubsection{Blastwind Dataset}
We evaluate the performance on the test set using the Exact Match and Levenshtein Edit Similarity metrics. These allow us to concretely compare our results with the results of the baseline models, which have results on other programming languages.

\subsubsection{HumanEval-Haskell Dataset} \label{approach:dataset-2}
We manually analyze the performance of our models. To prevent domination of the results by a few large functions with a lot of splits, a maximum of five splits per function are pseudo-randomly chosen from the manually introduced splits. Then, on a case-by-case basis, an overview of the performance is created with respect to each point of interest, such that common pitfalls can be detected (RQ2), e.g., the ability to predict list comprehensions.
This leads to a total of 603 samples from the HumanEval-Haskell dataset.

\subsubsection{Output}
The models are applied on the test sets using \textit{beam search}.
Beam search is a sampling technique that can result in better predictions in exchange for higher inference costs.
Instead of continuously sampling the most likely token (i.e., greedy prediction), beam search tracks the top-$k$ sequences at every inference step.
In the end, this results in $k$ unique sequences, of which the one with the highest probability is used.
Beam search is a common technique, that has also been widely used in seq2seq modelling~\cite{wiseman2016sequencetosequence}.
Additionally, since beam search is deterministic we can be sure that our observed metric values are easily reproducible.

\section{Experiments Setup}

In this section, the research questions are presented, after which the datasets, evaluation settings and metrics, and configuration and implementation details are discussed. 

\subsection{Research Questions}
In this work, we aim to answer the following Research Questions (RQs):

\begin{itemize}
    \item \textbf{RQ1: How well do code completion models perform on Haskell?}
        This research question aims to answer whether there are quantitative differences in code completion performance between Haskell and imperative languages.
        Nearly all literature focuses on imperative, non-functional languages, so it is unclear whether Haskell (and functional languages in general) are more difficult for code completion models.
        As the chosen code completion models are pre-trained on a range of imperative languages, we also aim to gain an insight into whether this knowledge transfers well to Haskell.
    
    \item \textbf{RQ2: What are the most common pitfalls for code completion on Haskell?}
        Having considered quantitative differences in code completion,
        we now aim to understand what aspects of Haskell are difficult to code completion models.
        This understanding is crucial to improving code completion models.
        
\end{itemize}

\subsection{Dataset}

We use two datasets for our experiments.
The first dataset, Blastwind, used for training, consists of permissively licensed Haskell function implementations.
This dataset is publicly accessible on HuggingFace and contains a total of 3.26 million functions.\footnote{\url{https://huggingface.co/datasets/blastwind/github-code-haskell-function}}
We create the second dataset ourselves by translating HumanEval~\cite{chen2021evaluating} from Python to Haskell,
leading to 164 Haskell functions based on HumanEval.
This dataset is used for our manual evaluation, where we determine common mistakes made by the code completion models.
None of the datasets overlap with the training data of the selected models, as these models were not trained on any Haskell code.
Both datasets consist of Haskell function implementations, with the primary difference being that HumanEval always contains large informative comments, while this is rarely the case in real code, as found in Blastwind.
This could have an effect on the efficacy of the models, as syntactical elements such as comments have been shown to have a positive impact on code completion performance~\cite{van2023enriching}.

\subsection{Evaluation Setting and Metrics}
\label{sec:evaluation-and-metrics}
We choose the same metrics that were used to evaluate UniXcoder and CodeGPT to compare code completion performance on Haskell to the performance on other programming languages
\cite{guo2022unixcoder,lu2021codexglue}.
This allows us to directly compare with their results for Python and Java.

The first metric, \textit{exact match}, compares whether the prediction and the ground truth are the exact same.
The second metric, \textit{edit similarity}, uses the Levenshtein distance between the ground truth and the prediction to compute how close the prediction is to the ground truth.
The distance is determined by the number of insertions, deletions, and substitutions required to make the target match the ground truth.
Then, the edit similarity is computed as shown in \autoref{eq:es}.

\begin{equation}
    ES(p, g) = 1 - \frac{Levenshtein(p, g)}{max(\lvert p \rvert, \lvert g \rvert)}
    \label{eq:es}
\end{equation}

The ground truth and prediction are trimmed (i.e., heading and trailing spaces are removed) and spacing is normalized (i.e., sequences of white-spaces of arbitrary length are replaced with a single space) before using them to compute metric values.
We present the metric values on a scale of 0 to 100.

\subsection{Configuration and Implementation Details}

We fine-tune UniXcoder using the default parameters, which were also used by~\citeauthor{guo2022unixcoder} to fine-tune their model on Python code.\footnote{\url{https://github.com/microsoft/CodeBERT/tree/0b522a6d7b2e25456e52b1c99a8e9cc6cd2aa6e0/UniXcoder/downstream-tasks/code-completion}}
Similarly, we use the default parameters as used by~\citeauthor{lu2021codexglue} for training CodeGPT.\footnote{\url{https://github.com/microsoft/CodeXGLUE/tree/main/Code-Code/CodeCompletion-line}}
For both models, we use a batch size of 2.
For inference, we apply beam search with a beam size of three for both UniXcoder and CodeGPT.
Additionally, we set the maximum number of tokens to predict to 128.
This is sufficient for line completion.

Fine-tuning and inference were performed on a server equipped with an NVIDIA Tesla V100S GPU.
The time required to complete fine-tuning is reported per model in \autoref{tab:finetune_times}.
The large discrepancy between the two models is caused by differences in the ways in which the models organize their training inputs.
CodeGPT batches as many inputs together during training, whereas UniXcoder uses padding tokens to fill any space left after an input.
This leads to CodeGPT needing fewer steps in total, reducing its training time.


\begin{table}[tb]
    \caption{Fine-tuning times.}
    \label{tab:finetune_times}
    \centering
    \begin{tabular}{lr}
        \toprule
        \textbf{Model} & \textbf{Time} \\
        \midrule
        UniXcoder & 19 hours \\
        \cdashline{1-2}\noalign{\vskip \belowrulesep}
        CodeGPT & 12 hours \\
        \bottomrule
    \end{tabular}
\end{table}

\begin{table}[tb]
    \caption{Inference times.}
    \label{tab:inference_times}
    \centering
    \begin{tabular}{llrr}
        \toprule
        \textbf{Model}              & \textbf{Variant}  & \textbf{Blastwind}      & 
        \textbf{HumanEval-Haskell} \\
        \midrule
        \multirow{2}{*}{UniXcoder}  & Base              & {20 h 26 min}           & {13 min}                      \\
                                    & Fine-tuned        & {1 h 50 min}            & {2 min}                       \\
        \cdashline{1-4}\noalign{\vskip \belowrulesep}
        \multirow{2}{*}{CodeGPT}    & Base              & {8 h 47 min}            & {3 min}                       \\
                                    & Fine-tuned        & {4 h 35 min}            & {4 min}                       \\
        \bottomrule
    \end{tabular}
\end{table}

The inference time for all models on Blastwind and HumanEval-Haskell are displayed in \autoref{tab:inference_times}.
The base models nearly always have a substantially higher processing time than the fine-tuned models, mainly due to the significantly longer predictions they produce.
These models have a weak understanding of Haskell, causing them to struggle with identifying where to end lines.
As a result, the inference times are increased due to the cost incurred by each generated token.
The difference is especially pronounced between the base and fine-tuned UniXcoder models.
This is to the fact that the base UniXcoder model was not trained to predict and stop at \textbf{<EOL>} tokens at the end of each line.
We post-process its outputs such that we only consider the first line it predicts, which ensures that this does not impact our results.
However, this does lead to higher inference times for the base model.

\section{Results}
\label{sec:Results}

We conduct a quantitative and qualitative analysis to determine how well UniXcoder and CodeGPT perform on Haskell.

\subsection{RQ1: How well do code completion models perform on Haskell?}

\begin{table}[tb]
    \caption{Blastwind \& HumanEval-Haskell Results. EM and ES expressed as number between 0 and 100.}
    \label{tab:dataset_results}
    \centering
    \begin{tabular}{llcccc}
        \toprule
        \multicolumn{2}{c}{\multirow{2}{*}{\textbf{Model}}}& \multicolumn{2}{c}{\textbf{Blastwind}} &
              \multicolumn{2}{c}{\textbf{HumanEval-Haskell}} \\
        \multicolumn{2}{c}{}&{\textbf{EM}} & {\textbf{ES}} & {\textbf{EM}} & {\textbf{ES}} \\

              \midrule
        \multirow{2}{*}{UniXcoder}  & Base        & {1.98}          & {25.93}                & {5.31}          & {27.31}         \\
                                    & Fine-tuned     & {28.00}         & {56.90}     & {13.10}         & {44.16}         \\
        \cdashline{1-6}\noalign{\vskip \belowrulesep}
        \multirow{2}{*}{CodeGPT}    & Base         & {2.05}          & {19.51}       & {5.80}          & {23.17}         \\
                                    & Fine-tuned    & {17.40}         & {46.95}     & {15.42}         & {40.01}         \\
        \bottomrule
    \end{tabular}
\end{table}

We quantitatively evaluate the performance on Blastwind and HumanEval-Haskell by running all base and fine-tuned models against the datasets.
Then, we compute the average EM and ES scores over all test samples.
The results are shown in \autoref{tab:dataset_results}.
Overall, the base models (i.e., (pre-)trained on six programming languages, not including Haskell) show significantly worse performance than the fine-tuned models.
On Blastwind, the two base models are roughly on-par, but the fine-tuned UniXcoder performs much better than the fine-tuned CodeGPT model across metrics.
There is not such a clear distinction on HumanEval-Haskell: CodeGPT scores better on Exact Match, whilst UniXcoder scores better on Edit Similarity.
Performance on Blastwind is substantially better than performance on HumanEval-Haskell, especially when considering Exact Match.
Nevertheless, there is a large overall improvement in model performance when the models are fine-tuned.

\subsection{RQ2: What are the most common pitfalls for code
completion on Haskell?} \label{sec:results:rq2}
We conduct a manual qualitative analysis on HumanEval-Haskell, to further analyse why UniXcoder outscores CodeGPT and the common pitfalls both models have with Haskell linecompletion. Note, only the \textit{fine-tuned} variants of the models are considered in this context. After splitting the data as described in \autoref{approach:dataset-2}, we obtained insights into the common pitfalls by annotating the performance of the splits based on several (sub)categories. By manual inspection of all the predictions, we also found some predictions to be `valid', viz. semantically equal, resulting in an updated performance overview as illustrated in \autoref{tab:dataset_2_results:manual-correction}. 

\begin{table}[h]
    \caption{Updated performance of fine-tuned models by manual inspection, where some non-EM predictions have been marked as `valid'. The correct ratio of predictions (\%), denoted by the `\%' header, is calculated by dividing the total count of predictions with the \textit{|EM| + Valid} count.}
    \label{tab:dataset_2_results:manual-correction}
    \centering
    \begin{tabular}{lccccc}
        \toprule
        \textbf{Model}              & \textbf{|EM|} & \textbf{Valid} & \textbf{|EM| + Valid} & \textbf{Total} & \textbf{\%}\\
        \midrule
        UniXcoder & 79 & 18 & 97 & 603 & 16.09 \\
        CodeGPT & 93 & 20 & 113 & 603 & 18.74 \\
        \bottomrule
    \end{tabular}
\end{table}

\noindent When looking at the distribution of annotations for both CodeGPT and UniXcoder (see Appendix C~\cite{authors2024replication}), there is no substantial dissimilarity in their general performance with regard to the prediction of certain Haskell (sub)categories such as if/then/else statements, generators, guards, functions, lists, logical operators, arithmetic operators, and case expressions. However, a clear difference in their general performance is shown by the distribution of the other annotations listed in \autoref{tab:codegpt_vs_unixcoder:distribution_annotations}. This table shows that CodeGPT has significantly more \texttt{empty} predictions when compared to UniXcoder, however, UniXcoder has more \texttt{incomplete}, \texttt{wrong syntax}, and \texttt{undefined} as predictions. Furthermore, a few specific predictions of UniXcoder have been separately marked as worth mentioning. This includes:
\begin{itemize}
    \item using a variable out of scope, mentioned in a previous function within the provided context;
    \item using a variable out of scope, mentioned in a comment within the provided context, yet not defined within the Haskell let context;
    \item line completion of max characters due to getting stuck in a repetitive loop of predicting values for a list.
\end{itemize}

\begin{table}[h]
    \caption{Distribution of distinctive annotations between CodeGPT and UniXcoder for predictions that were neither an exact match nor deemed valid.}
    \label{tab:codegpt_vs_unixcoder:distribution_annotations}
    \centering
    \begin{tabular}{lccccc}
        \toprule
        \textbf{Annotation}              & \textbf{CodeGPT} & \textbf{UniXcoder} \\
        \midrule
        Wrong type & 12 & 26 \\
        Wrong value & 69 & 66 \\
        Wrong function & 108 & 81 \\
        Empty prediction & 106 & 1 \\
        Incomplete prediction & 30 & 130 \\
        Wrong syntax & 13 & 96 \\
        `\texttt{undefined}' keyword & 1 & 31 \\
        \bottomrule
    \end{tabular}
\end{table}

\noindent As there seemed to be a correlated overlap in certain annotations for each specific prediction done by UniXcoder and CodeGPT, the commonalities in annotations have also been researched (see Appendix A\footnotemark[9]{}) in order to get more insight into the different behavior of UniXcoder and CodeGPT. In addition, the commonalities between CodeGPT's and UniXcoder's annotations per prediction itself, hereafter referred to as \textit{overlaps}, have been researched to get more context for the identification of common pitfalls. The most insightful annotation links that have been discovered, e.g. high overlap or similar annotation type, are illustrated in \autoref{tab:manual-eval:most-insightful-links}. One annotation in particular, i.e., \texttt{extra comment}, has been analyzed separately in Appendix B\footnotemark[10]{}, as CodeGPT adds an extra comment to a lot of its predictions, which was found to often be of similar syntax. There is no clear pattern to where CodeGPT adds these comments 
, however, its content was always in the following format: "\textit{| Creates a value of `<some class name starting with ProjectsLocations>’
with the minimum fields required to make a request.}".

\begin{table}[h]
    \caption{Most insightful overlaps between annotations of predictions. CodeGPT is denoted by \texttt{C} and UniXcoder by \texttt{U}.}
    \label{tab:manual-eval:most-insightful-links}
    \centering
    \begin{tabular}{lll}
        \toprule
        \textbf{Annotation}              & \textbf{Annotation}      & \textbf{Overlap} \\
        \midrule
        \texttt{C}: empty                   & \texttt{U}: incomplete            & 54.72\% (58/106)        \\
        \texttt{C}: empty                   & \texttt{U}: undefined             & 14.15\% (15/106)        \\
        \texttt{C}: incomplete              & \texttt{U}: incomplete            & 43.33\% (13/30)        \\
        \midrule
        \texttt{U}: undefined              & \texttt{C}: empty            & 48.39\% (15/31)        \\
        \texttt{U}: incomplete              & \texttt{C}: empty            & 44.62\% (58/130)        \\
        \midrule
        \texttt{C}: complete function              & \texttt{C}: wrong function            & 75.00\% (9/12)        \\
        \texttt{C}: variable definition              & \texttt{C}: wrong value            & 28.57\% (8/28)        \\
        \texttt{C}: valid              & \texttt{C}: extra comment            & 15.04\% (17/113)        \\
        \texttt{C}: arithmetic logic              & \texttt{C}: wrong value            & 14.29\% (12/84)        \\
        \midrule
        \texttt{U}: undefined & \texttt{U}: case expr. (body) & 67.74\% (21/31) \\
        \texttt{U}: incomplete & \texttt{U}: case expr. (body) & 58.46\% (76/130) \\
        \texttt{U}: wrong type & \texttt{U}: case expr. (body) & 38.46\% (10/26) \\
        \texttt{U}: wrong function & \texttt{U}: case expr. (body) & 38.27\% (31/81) \\
        \texttt{U}: variable definition & \texttt{U}: incomplete & 31.58\% (6/19 \\
        \texttt{U}: arithmetic logic              & \texttt{U}: wrong value            & 17.24\% (15/87)        \\
        \bottomrule
    \end{tabular}
\end{table}

\section{Discussion}
This section discusses the results and what they imply. Also the validity and what conclusions can be drawn from the results.

\subsection{RQ1: How well do code completion models perform on Haskell?}

\begin{table}[tb]
    \caption{PY150 and JavaCorpus results.}
    \label{tab:unixcoder_codegpt_paper_results}
    \centering
    \begin{tabular}{llrr}
        \toprule
        \textbf{Dataset}            & \textbf{Model}    & \textbf{EM}   & \textbf{ES}   \\
        \midrule
        \multirow{2}{*}{PY150}      & UniXcoder         & 43.12         & 72.00         \\
                                    & CodeGPT           & 39.11         & 69.69         \\
        \cdashline{1-4}\noalign{\vskip \belowrulesep}
        \multirow{2}{*}{JavaCorpus} & UniXcoder         & 32.90         & 65.78         \\
                                    & CodeGPT           & 25.30         & 61.54         \\
        \bottomrule
    \end{tabular}
\end{table}

The performance of UniXcoder and CodeGPT on Haskell improved drastically after fine-tuning.
When considering the Blastwind dataset, our fine-tuned models exhibit worse performance compared to the models when fine-tuned on Python and Java~\cite{guo2022unixcoder,lu2021codexglue}, as displayed in \autoref{tab:unixcoder_codegpt_paper_results}.
The relative improvements to the base models affirm that language models can become sufficient in Haskell, despite its stark differences to typical programming languages.
The poor performance relative to Python and Java could indicate differences in difficulty between the languages, but may also be explained by differences in dataset size and quality.
Our Haskell train set consists of \numprint{30349824} tokens, whilst PY150~\cite{raychev2016probabilistic} contains \numprint{154241924} tokens for training, and JavaCorpus~\cite{allamanis2013mining} contains \numprint{24029629} tokens for training.\footnote{All tokens counts are using the UniXcoder tokenizer}
Based on the number of tokens used for training and the observed metric values,
we believe that Haskell is in fact more challenging than Python and Java overall.
Nevertheless, further investigations with larger high-quality Haskell training sets (both in terms of tokens and in terms of prompt length) are necessary to validate these claims.

The results for the HumanEval-Haskell dataset are different from the results on the Blastwind dataset.
Overall, the base models score better on this dataset, whereas both fine-tuned models score substantially lower.
The superior performance of the base models could be explained by the format and context of HumanEval-Haskell.
First off, each sample contains a relatively large informative comment.
As both UniXcoder and CodeGPT are pre-trained on a large corpus of natural language, this naturally enhances their ability to infer the appropriate following tokens.
In contrast, the Blastwind dataset contains smaller comments, leaving out details that are unimportant or obvious to humans.
Secondly, samples in the HumanEval-Haskell dataset are far more self-contained than the samples in the Blastwind dataset.
Samples in HumanEval-Haskell are typically single functions, but may also have numerous helper functions located in the same file.
Functions in the Blastwind dataset are not guaranteed to be self-contained.
The samples in this dataset are solely individual function implementations, without including any external helper functions that the model may need.
This subsequently makes it more difficult for the models to accurately predict the following code: helper functions, despite being in scope, are not known to the model.
The fine-tuned models may suffer from this to a lesser extent due to the training data resembling the test set more.

Despite the HumanEval-Haskell dataset including more informative comments, the fine-tuned models perform better on the Blastwind dataset.
This could similarly be explained by the dissimilarity between the data in the two datasets: the fine-tuned models were trained on the Blastwind dataset, and therefore are trained to use little additional data when predicting Haskell.
While the models have natural language understanding, they have not been trained to relate natural language to Haskell code.
This could explain the difficulty in using the information embedded in the informative comments.
Training approaches data that combines natural language with code, such as commented functions, or Q\&A-style conversations about code (as seen on StackOverflow) could alleviate this.

Overall, for both datasets the fine-tuned models outperform the base models by a substantial margin, indicating that fine-tuning code completion models on Haskell is crucial to achieving optimal performance.
Our results indicate that Haskell is harder to predict than Python and Java, but further experiments are required to verify this.


\subsection{RQ2: What are the most common pitfalls for code completion on Haskell?} \label{sec:discussion:rq2}

The distribution of the annotations for the HumanEval-Haskell predictions in combination with the general performance indicates a difference in the capabilities of the models. The distinctive factor in the capabilities can be explained by their differences in behavior. While CodeGPT is more cautious as evidenced by often predicting empty lines, UniXcoder tends to not only predict \texttt{undefined} in such cases but also shows a more aggressive prediction behavior. This behavior of UniXcoder leads to incomplete predictions (also often wrong functions in such cases) that would require a new line to complete in order to successfully continue. Furthermore, UniXcoder spits out a troublesome number of predictions featuring fundamental issues such as wrong syntax (e.g., mismatched brackets or capitalized function names), scope issues, or getting stuck in repetition. Ultimately, this makes CodeGPT a safer choice for practical usage, purely based on the behavior of the models. In addition, CodeGPT demonstrated to be the better choice after manually evaluating the correctness of HumanEval-Haskell predictions.
It is worth mentioning that splits are introduced manually on points of interest (based on developer experience) for the HumanEval-Haskell dataset, while for the Blastwind dataset, the splits are introduced pseudo-randomly. This variation in the method might well influence the resulting predictions, as the manually introduced splits could be above averagely complex for the models to predict. The exact influence of this variation remains unclear, but it should be noted that the results of this deviation are included in the EM and ES values.
Regarding the common pitfalls of Haskell code completion, none of the annotated categories in the manual evaluation yield a significant performance disparity. In fact, the overall performance indicates a high demand for improved support to excel in any particular category, which should be the primary focus of future research.

\subsection{Implications}
The results of our process of fine-tuning LLMs such as CodeGPT and UniXcoder on Haskell datasets have several theoretical and practical implications for the field of functional programming and code-LLMs. 

\subsubsection{Theoretical Implications}
Fine-tuning LLMs on Haskell has shown to be effective for these models to grasp functional programming concepts.
Furthermore, the nuanced trade-off between empty and faulty completions during manual evaluations underlines the complexity of optimal decision-making in AI, suggesting the need for more advanced metrics in model performance assessment.
Additionally, the fact that multilingual LLMs underperform significantly compared to their fine-tuned counterparts shows the need for diverse high-quality Haskell datasets in the pre-training of these LLMs.
Adding a deep understanding of functional programming might elevate the predictions for OOP-based languages as well, but further research is needed to accurately determine the effect of functional programming pre-training on OOP languages. 
These datasets could then be included in the pre-training of LLMs for a thorough understanding of this functional programming language.

\subsubsection{Practical Implications}
Practically, the fine-tuned LLMs on Haskell show to be promising for sophisticated developer tools in functional programming, for instance for code search, code repair, or code summarization.
These models could serve as a helper tool for developers new to the complex world of functional programming.
For instance, insights gained from AI interactions with Haskell might inspire new language features or paradigms.
Furthermore, these models could enable more advanced code suggestion and debugging features, significantly reducing the time and effort required for developers to write and maintain Haskell code.
Training LLMs on Haskell can also be beneficial for a deeper understanding of OOP languages that implement functional programming constructs, such as higher-order functions, the notion of pure functions, and recursion.

\subsection{Future Work}
Future work could strengthen our findings by repeating our experiments on other models, other functional languages, or other datasets.
Constructing new high-quality Haskell datasets would be particularly interesting for future LM-based tools for Haskell.
At present, there are no high-quality curated Haskell datasets, which can lead to sub-par models resulting from issues such as duplication, small dataset size, and uninformative samples.
The fact that our filtering process eliminated nearly 90\% of samples further emphasizes the necessity of high-quality Haskell datasets.
Moreover, the significant number of discarded samples suggests that conducting experiments with larger datasets could yield valuable insights.
Alternatively, online data from e.g., StackOverflow could be used to create an understanding of natural language when related to Haskell code.
Other research could also investigate the effect of including Haskell datasets in its pre-training steps on mainstream languages used for evaluating LLMs.


Furthermore, it is worth considering alternative training inputs.
Our chosen dataset only includes one function implementation per sample.
However, utilizing complete files as training inputs may yield more precise results.
Function implementation often depends on contextual information located in the same file.
Hence, training on full files instead of single-function implementations would also make our training data align more closely with real-life scenarios.

Finally, determining whether understanding of different programming languages transfers to Haskell would provide an interesting insight.
In this study, the pre-trained models were trained on six diverse programming languages before being fine-tuned for predicting Haskell code.
The underlying assumption was that this wide-ranging knowledge base would yield better results in Haskell.
However, determining whether this approach was indeed beneficial is beyond the scope of this paper.
Hence, future research could explore whether this assumption is correct by comparing our findings with results from models that were solely trained on natural language or a single programming language.


\subsection{Threats to the Validity}
Due to the experimental method of this paper, the validity of the results in the context of the `real world' must also be considered. These threads can be divided into three categories: threats to internal validity, external validity, and construct validity.

\subsubsection{Internal Validity} 
This section discusses the elements that impact the model's performance, external factors that accidentally affect the outcomes, and mistakes made during the implementation process.
Translation quality in the HumanEval-Haskell dataset is important because translation errors could lead to incorrect measurements of performance, compromising our ability to gauge which aspects of Haskell are most difficult to language models. 
Additionally, any data leakage between train and test sets must be mitigated to prevent unrealistic perceptions of performance.

We have considered these factors and implemented steps to minimize or mitigate the effects of these threats.
For instance, the translated HumanEval functions were reviewed by numerous authors,
and the Blastwind dataset was deduplicated to promote diversity in the training data as well as to prevent leakage between train and test sets.
Moreover, we split the Blastwind dataset into train and test sets on a repository basis, meaning that all code from the same repository will be in either the train or the test set, but never both.
This method further serves to prevent data leakage.
To allow further examination regarding these concerns, we have made all resources, including datasets and models, publicly available for scrutiny and further research.

\subsubsection{External Validity}
This section discusses the elements that may influence the applicability or broader relevance of our research results. 
The usage of line completion in a realistic programming environment may differ from the experimental setup designed to answer \textit{RQ1}.
We introduce splits in pseudo-random locations, given that there is a sufficient number of surrounding tokens.
A programmer using the line completion functionality would likely invoke code completion in different places, such as trigger points.
Examples of trigger points are property accesses on variables (e.g., a \texttt{.}, or \texttt{->}), or for example an opening bracket (\texttt{(}).
This limitation is, however, partly mitigated in \textit{RQ2}, in which we manually introduce the splits in logical places. To prevent skewed results, the datasets are deduplicated. There is no overlap in pre-train data and test data, as the pre-train data for both UniXcoder and CodeGPT do not include Haskell code~\cite{lu2021codexglue, guo2022unixcoder, husain2019codesearchnet}.

In the case of HumanEval-Haskell, the model input includes a detailed comment on the functionality of the desired function and example input and output. In reality, however, this context will often not be available for the model, which means that the prediction has to be made using less context, resulting in worse performance.

Furthermore, the setting of this evaluation is relatively pure in HumanEval-Haskell, each time the full comment is available, no surrounding other functions (which could influence the model's predictions), and it only has to predict a single line. In a real-world setting, such as code completions within a developer's project in an IDE, a lot more context is presented which might distract the model from giving the proper completions.

Additionally, It is essential that the Blastwind dataset broadly represents Haskell code patterns to prevent training biases. 
Insufficient diversity can lead to a model performing poorly on real-world Haskell problems.

\subsubsection{Construct Validity}
This section discusses the validity of the measurements performed. 
Metrics must accurately reflect the model's ability to generate functional Haskell code -- misaligned evaluation can invalidate the perceived effectiveness of the fine-tuning process.
As described in \autoref{sec:evaluation-and-metrics}, the metrics used are commonly used in literature~\cite{svyatkovskiy2020intellicode,guo2022unixcoder,lu2021codexglue}.
While it is known that these metrics do not capture all nuances of code, such as semantics, they are still widely used and thus essential to be able to compare models with one another.
Hence, the chosen combination of metrics ensures a sufficient evaluation of the output of the model. 
Additionally, small differences in interpretation of the metrics can result in different conclusions. It is therefore essential to properly elaborate on the usage and interpretation of the metrics, as done in \autoref{sec:evaluation-and-metrics} and \autoref{sec:Results}. The same holds for the process of data processing, for which a detailed description is given in \autoref{sec:Approach}.

\section{Conclusion}

LLMs for code completion are often trained only on imperative or Object-Oriented Programming languages.
As functional languages are severely underrepresented in training data, the performance of code completion on these languages is substantially worse than on other languages.
In this work, we explore the performance of two multilingual auto-regressive language models, UniXcoder and CodeGPT, when tasked to perform line completion on Haskell function implementations.
Results show that both models perform significantly better after being fine-tuned, indicating that knowledge of imperative languages does not necessarily transfer to functional languages.
Base models perform considerably worse, suggesting that prior knowledge of imperative programming languages may not transfer well to functional languages, and indicating a need for future datasets to include high-quality Haskell code.
Additionally, our manual analysis shows that CodeGPT tends to generate more empty predictions and unnecessary comments, while UniXcoder generates incomplete, wrong syntax, and `undefined' predictions. Since the behavior of UniXcoder shows more fundamental completion issues than CodeGPT during manual evaluation, CodeGPT is a safer choice for practical usage.
Regarding the primary pitfalls of Haskell code completion, no specific categories annotated in the manual evaluation indicate a substantial difference in performance such that a general statement can be made about pivotal focus areas for the improvement of Haskell code completion -- the language in general requires more support to perform well in any category as of yet.
Lastly, the community should take into consideration the implementation of FP languages when training LLMs, especially since many modern OOP languages are integrating more functional concepts as time goes on.
Giving LLMs a wider understanding of these fundamental concepts may help models understand such concepts across many different languages.

\section{Data Availability}
To ensure reproducibility and replicability, we publish all data used for our training and evaluation, including all source code used to conduct our evaluation and analysis.
We publish our manually translated Haskell-HumanEval dataset, and
additionally provide a supplemental detailing overlapping annotations between UniXcoder and CodeGPT, and highlighting the extra comments predicted by CodeGPT.


\newpage
\bibliographystyle{ACM-Reference-Format}
\bibliography{main}

\end{document}